# LLM-Augmented Traffic Signal Control with LSTM-Based Traffic State Prediction and Safety-Constrained Decision Support

Jiazhao Shi[*a]
[a]Tandon School of Engineering, New York University, 6 MetroTech Center, Brooklyn, NY 11201, USA;
[*]js12624@nyu.edu

## ABSTRACT

Traffic signal control is a critical task in intelligent transportation systems, yet conventional fixed-time and rule-based methods often struggle to adapt to dynamic traffic demand and provide limited decision interpretability. This study proposes an LLM-augmented traffic signal control framework that integrates LSTM-based short-term traffic state prediction, predictive phase selection, structured large language model reasoning, and safety-constrained action filtering. The LSTM module forecasts future queue length, waiting time, vehicle count, and lane occupancy based on recent intersection-level observations. A predictive controller then generates candidate signal actions, while the LLM module evaluates these actions using structured traffic-state inputs and produces congestion diagnoses, phase adjustment recommendations, and natural-language explanations. To ensure operational reliability, all LLM-generated recommendations are validated by a safety filter before execution. Simulation-based experiments in SUMO compare the proposed method with fixed-time control, rule-based control, and an LSTM-based predictive baseline under balanced demand, directional peak demand, and sudden surge scenarios. The results indicate that the proposed framework improves traffic efficiency, especially under dynamic and non-recurrent traffic conditions, while maintaining zero constraint violations after safety filtering. Overall, this study demonstrates that LLMs can enhance traffic signal control when used as constrained reasoning and decision-support modules rather than direct low-level controllers.

## 1. INTRODUCTION

Traffic signal control is a fundamental component of intelligent transportation systems, directly affecting urban mobility, congestion mitigation, energy consumption, and road network efficiency. As urban traffic demand continues to grow, conventional fixed-time and actuated signal control methods often struggle to adapt to rapid fluctuations in traffic flow, directional imbalance, spillback, and non-recurrent congestion. Although rule-based and transportation-engineering approaches remain widely used because of their simplicity and interpretability, their ability to respond to complex spatiotemporal traffic dynamics is limited, especially when traffic patterns change over time or when local intersection decisions influence upstream and downstream network states.

In recent years, data-driven traffic signal control has attracted increasing attention. Reinforcement learning methods have been applied to adaptive traffic signal control by modeling each intersection as an agent that selects signal phases based on queue length, waiting time, pressure, or vehicle delay [1][2]. Representative studies such as IntelliLight, PressLight, CoLight, and AttendLight have shown that learning-based controllers can outperform fixed-time and manually designed policies under many simulated traffic conditions [3][4]. However, these methods still face several challenges. First, reinforcement learning controllers often require extensive training and may generalize poorly across unseen road networks, traffic demand distributions, and incident scenarios. Second, their decision-making processes are usually difficult for human traffic managers to interpret. Third, purely reactive controllers may fail to anticipate near-future congestion propagation if they rely mainly on current traffic states.

Traffic state prediction provides an important complementary direction for adaptive signal control. Long short-term memory networks were originally proposed to capture long-range temporal dependencies in sequential data [5], and they have been widely used in traffic forecasting because traffic flow, speed, occupancy, and queue length naturally evolve over time. More recent spatiotemporal forecasting methods further combine recurrent networks, graph neural networks, attention mechanisms, and state-space models to capture both temporal evolution and spatial dependency over road networks. For example, DCRNN models traffic forecasting as a diffusion process on directed graphs [6], STGCN introduces graph convolution and temporal convolution for structured traffic sequences [7], Graph WaveNet learns adaptive graph structures for multivariate time-series forecasting [8], and GMAN applies attention mechanisms to model spatiotemporal correlations [9]. More recently, FAST integrates attention and state-space models to achieve scalable spatiotemporal traffic prediction over large sensor networks [13]. These studies suggest that accurate short-term traffic prediction can provide valuable foresight for signal control. Nevertheless, prediction alone does not automatically translate into interpretable or safety-aware control decisions.

Large language models provide a new opportunity to bridge this gap. Unlike conventional neural predictors, LLMs are capable of processing structured textual descriptions, reasoning over multi-source information, generating human-readable explanations, and following explicit operational constraints. Recent studies have begun to explore the use of LLMs in traffic signal control and urban mobility systems. LLMLight investigates the potential of LLMs as traffic signal control agents, highlighting their reasoning, generalization, and interpretability capabilities [10]. CoLLMLight extends this idea toward cooperative LLM agents for network-wide traffic signal control [11]. In parallel, LLM-based and generative urban intelligence methods are increasingly being explored for simulation, scene understanding, and decision support. For example, ChatSUMO uses large language models to automate traffic scenario generation in SUMO [12], while recent neural scene and driving world model studies, including ProSGNeRF and GaussianDWM, demonstrate the broader trend of using generative and language-aligned models for urban scene understanding and driving-environment representation [14], [15].

Despite these advances, directly using an LLM as a real-time traffic signal controller remains problematic. Traffic signal control is a safety-critical and time-sensitive task. Signal actions must satisfy strict operational constraints, including minimum green time, maximum green time, yellow intervals, pedestrian crossing requirements, and emergency vehicle priority. LLM outputs may be inconsistent if not properly constrained, and their latency may be unsuitable for high-frequency control. Therefore, a more practical design is to use the LLM not as a low-level controller, but as an upper-level reasoning and decision-support module. Under this design, a conventional or learning-based controller can generate candidate signal actions, while the LLM interprets predicted traffic conditions, diagnoses abnormal congestion patterns, recommends parameter adjustments, and provides explanations that are understandable to human operators.

Motivated by this perspective, this study proposes an LLM-augmented traffic signal control framework that integrates LSTM-based traffic state prediction with structured LLM reasoning and safety-constrained decision support. The proposed framework first collects intersection-level traffic states from a microscopic simulation environment, including queue length, waiting time, vehicle count, average speed, lane occupancy, and current signal phase. An LSTM module then predicts short-term future traffic states based on recent observations. A predictive signal controller uses these forecasts to generate candidate phase decisions, such as extending the current phase or switching to a competing phase. Instead of directly replacing this controller, the LLM module receives a structured summary of the current state, predicted state, candidate action, and operational constraints. It then produces a constrained recommendation, including whether to accept the candidate action, whether to adjust the phase duration, whether abnormal congestion is present, and why the decision is reasonable. Finally, a safety filter validates the recommended action before execution.

This design has three advantages. First, it combines the temporal modeling ability of LSTM with the semantic reasoning ability of LLMs. The LSTM baseline provides quantitative short-term forecasts, while the LLM module converts these forecasts into interpretable decision support. Second, the framework preserves operational safety by ensuring that all final actions pass through explicit traffic-signal constraints. Third, the method produces both control actions and natural-language explanations, making it more transparent than black-box reinforcement learning controllers. This is particularly important for real-world traffic management, where human operators often need to understand why a signal plan was modified before trusting an automated system.

The main contributions of this paper are summarized as follows:

1. We propose an LLM-augmented traffic signal control framework that combines LSTM-based short-term traffic state prediction, predictive phase selection, structured LLM reasoning, and safety-constrained action filtering.
2. We design an LSTM-based predictive control baseline that forecasts future queue length, waiting time, and traffic demand, and then selects signal actions based on predicted phase pressure or demand score.
3. We introduce a structured LLM decision-support module that generates interpretable recommendations, congestion diagnosis, risk flags, and safety-aware explanations without directly replacing the low-level traffic controller.
4. We evaluate the proposed framework in a SUMO simulation environment under multiple demand scenarios, including balanced traffic, directional peak demand, and sudden traffic surge, using metrics such as average waiting time, queue length, travel time, throughput, number of stops, and emission-related indicators.

By positioning the LLM as a reasoning and decision-support layer rather than an unconstrained real-time controller, this study aims to provide a practical and interpretable path toward next-generation intelligent traffic signal control.

## 2. METHODS

**2.1. Overview of the Proposed Framework**

This study proposes an LLM-augmented traffic signal control framework that combines short-term traffic state prediction, predictive signal phase selection, structured LLM-based decision support, and safety-constrained action filtering. The overall objective is to improve adaptive traffic signal control under dynamic traffic demand while enhancing the interpretability and operational safety of signal decisions.

The proposed framework consists of four major components:
1. a traffic state representation module,
2. an LSTM-based traffic state prediction module,
3. a predictive signal control module, and
4. an LLM-augmented decision-support module with a safety filter.

At each decision step, traffic states are collected from the intersection, including queue length, waiting time, vehicle count, average speed, lane occupancy, and the current signal phase. The LSTM module uses recent historical observations to forecast near-future traffic conditions. Based on the predicted traffic state, a predictive controller generates a candidate signal action, such as extending the current green phase or switching to another phase. The LLM module then receives a structured input containing the current state, predicted traffic state, candidate action, and signal operation constraints. It produces an interpretable recommendation and explanation. Finally, a safety filter checks whether the recommended action satisfies traffic signal constraints before the final action is executed in the simulation environment.

The framework is designed so that the LLM does not directly replace the low-level signal controller. Instead, it serves as a higher-level reasoning and decision-support module. This design reduces the risk of unconstrained or unstable LLM-generated control actions while preserving the interpretability advantage of large language models.

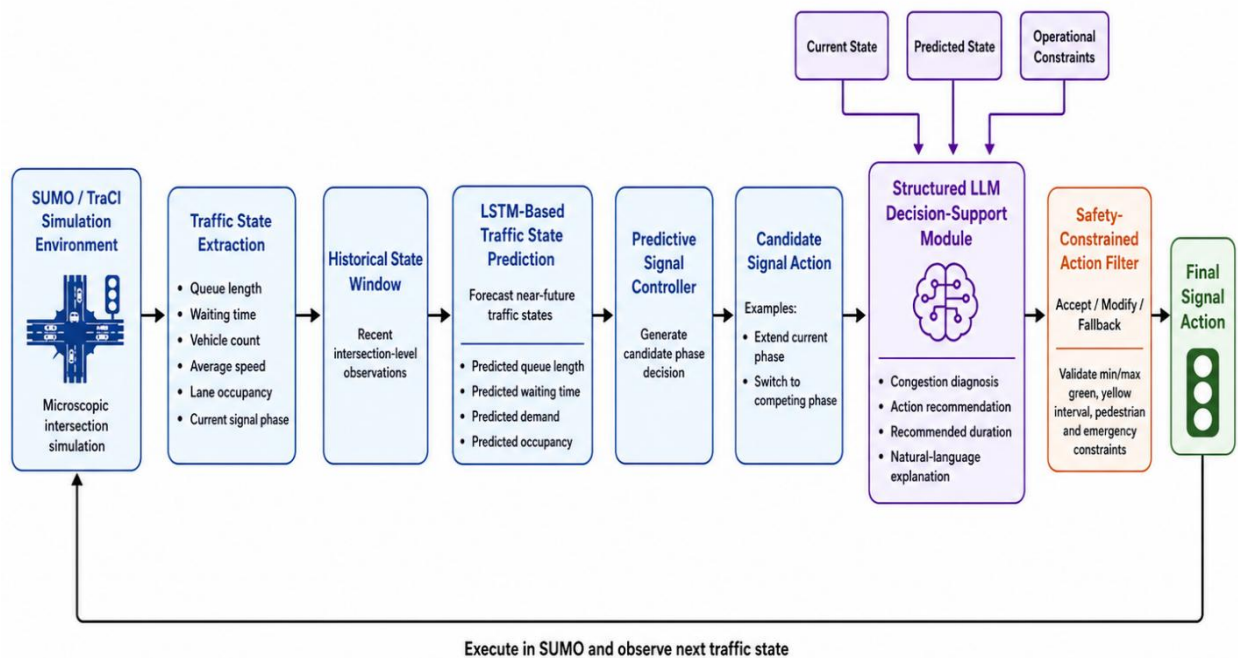

Figure 1. Overall architecture of the proposed LLM-augmented traffic signal control framework

## 2.2. Problem Formulation

We formulate adaptive traffic signal control as a sequential decision-making problem. Consider an intersection with a set of incoming lanes $L$ and a set of feasible signal phases $A$. At each control step $t$, the controller observes the current traffic state $s_t$, selects a signal action $a_t \in A$, and applies the action for a fixed or adaptive duration.

The traffic state at time $t$ is represented as:
$$s_t = [q_t, w_t, n_t, v_t, o_t, p_t],$$

where $q_t$ denotes queue length, $w_t$ denotes accumulated waiting time, $n_t$ denotes vehicle count, $v_t$ denotes average vehicle speed, $o_t$ denotes lane occupancy, and $p_t$ denotes the current signal phase. These variables are collected for each incoming lane or approach and then organized into an intersection-level state vector.

Given a historical observation window of length $K$, the input sequence is defined as:
$$X_t = [s_{t-K+1}, s_{t-K+2}, \ldots, s_t].$$

The objective of the prediction module is to estimate future traffic states over a prediction horizon $H$:
$$\hat{Y}_{t+1:t+H} = f_\theta(X_t),$$

where $f_\theta$ is the LSTM-based forecasting model and $\hat{Y}_{t+1:t+H}$ includes predicted queue length, waiting time, traffic demand, and occupancy.

The signal control objective is to minimize traffic inefficiency over the simulation period. The primary evaluation metrics include average waiting time, average queue length, average travel time, number of stops, throughput, and emission-related indicators. In the control process, the controller seeks to select actions that reduce future congestion while satisfying all signal timing constraints.

## 2.3. Traffic State Representation

The traffic state representation is designed to capture both current congestion conditions and short-term temporal evolution. For each incoming lane or movement, we extract the following features:
$$x_t^l = [q_t^l, w_t^l, n_t^l, v_t^l, o_t^l],$$

where $l \in L$ denotes a lane or approach. Specifically, $q_t^l$ is the number of halted vehicles on lane $l$, $w_t^l$ is the accumulated waiting time, $n_t^l$ is the number of vehicles detected within the lane or approach, $v_t^l$ is the average speed, and $o_t^l$ is the occupancy rate.

The intersection-level traffic state is obtained by concatenating the lane-level features and adding the current signal phase representation:
$$s_t = [x_t^1, x_t^2, \ldots, x_t^{|L|}, p_t].$$

The current signal phase $p_t$ is encoded as a one-hot vector. For example, in a two-phase intersection, the phase can be encoded as:
$$p_t = \begin{cases} [1,0], & \text{if the north-south direction has green light,} \\ [0,1], & \text{if the east-west direction has green light.} \end{cases}$$

This representation allows the prediction model to learn how the traffic state evolves under different signal phases.

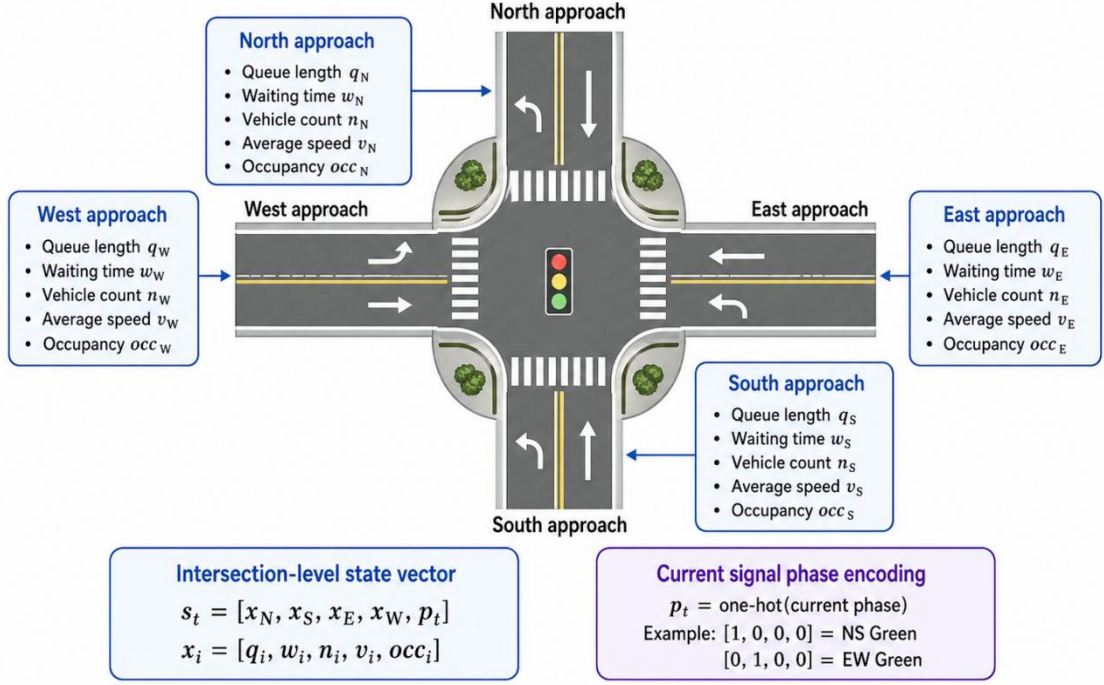

Figure 2. Traffic state representation at a four-arm intersection

### 2.4. LSTM-Based Traffic State Prediction

The LSTM module is used to model the temporal dependency of traffic states. Traffic conditions at an intersection are inherently sequential: queue length, waiting time, and lane occupancy are not independent at each time step, but are strongly influenced by previous signal phases and historical traffic demand. Therefore, LSTM is selected as the primary temporal prediction baseline because of its ability to capture long-term dependencies in sequential data.

Given the historical traffic state sequence $X_t$, the LSTM updates its hidden state as follows:
$$h_t = \text{LSTM}(s_t, h_{t-1}),$$

where $h_t$ is the hidden representation at time $t$. The final hidden state is passed through a fully connected output layer to predict the future traffic state:
$$\hat{Y}_{t+1:t+H} = W_o h_t + b_o,$$

where $W_o$ and $b_o$ are learnable parameters.

The prediction target includes future queue length and waiting time for each approach:
$$\hat{Y}_{t+1:t+H} = [\hat{q}_{t+1:t+H}, \hat{w}_{t+1:t+H}, \hat{n}_{t+1:t+H}, \hat{o}_{t+1:t+H}].$$

The LSTM model is trained by minimizing the mean squared error between the predicted and observed future traffic states:
$$\mathcal{L}_{pred} = \frac{1}{N} \sum_{i=1}^{N} \| Y_i - \hat{Y}_i \|_2^2,$$

where $N$ is the number of training samples, $Y_i$ is the ground-truth future traffic state, and $\hat{Y}_i$ is the predicted traffic state. In this study, the LSTM prediction model serves two roles. First, it provides a strong temporal forecasting baseline. Second, it provides future traffic estimates for the proposed LLM-augmented control framework.

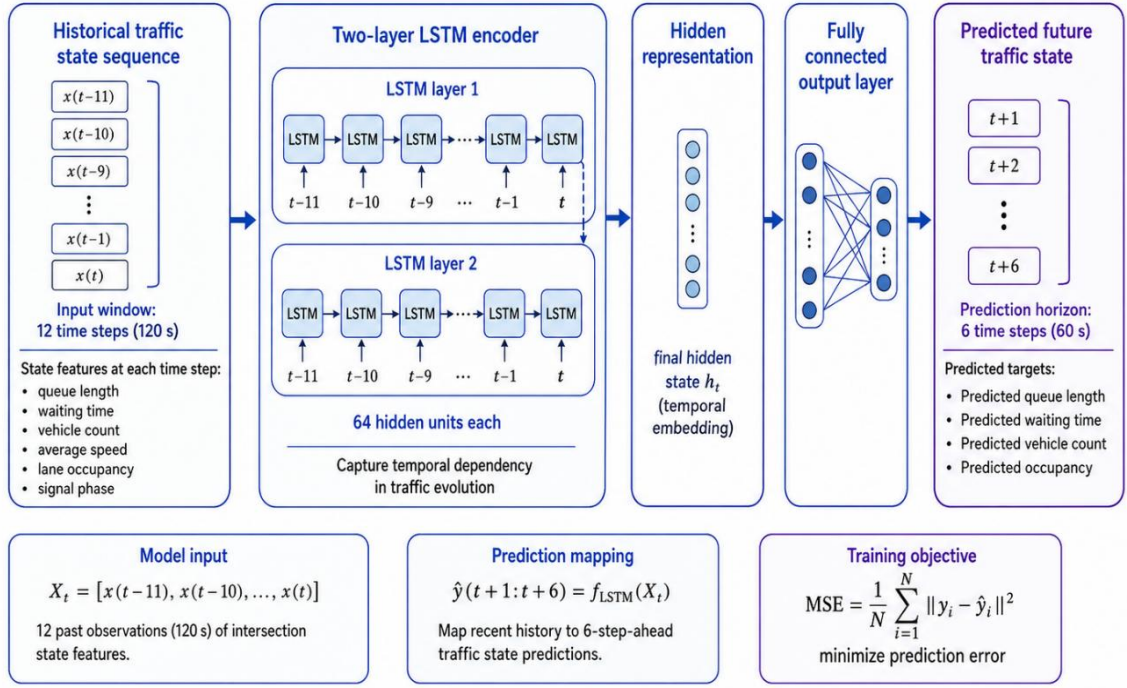

Figure 3. LSTM-based short-term traffic state prediction module

### 2.5. LSTM-Based Predictive Signal Control Baseline

To evaluate the contribution of the LLM module, we construct an LSTM-based predictive control baseline. This baseline uses the predicted future traffic state to select the next signal phase without LLM assistance.

For each candidate signal phase $a \in A$, we calculate a predicted demand score based on future queue length, waiting time, and lane occupancy:

$$D(a) = \sum_{l \in L(a)} (\alpha \hat{q}^l_{t+1:t+H} + \beta \hat{w}^l_{t+1:t+H} + \gamma \hat{o}^l_{t+1:t+H}),$$

where $L(a)$ is the set of lanes served by phase $a$, and $\alpha$, $\beta$, and $\gamma$ are weighting coefficients for queue length, waiting time, and occupancy, respectively.

The baseline controller selects the phase with the highest predicted demand score:

$$a_t^{LSTM} = \arg \max_{a \in A} D(a).$$

If the selected phase is the same as the current phase, the controller extends the current green phase. Otherwise, the controller switches to the selected phase after inserting the required yellow interval. This LSTM-based predictive controller provides a direct comparison against the proposed LLM-augmented method. It allows us to evaluate whether the LLM module provides additional benefit beyond short-term traffic prediction alone.

### 2.6. LLM-Augmented Decision-Support Module

The LLM-augmented decision-support module is designed to enhance the interpretability and adaptability of the predictive controller. Instead of directly generating signal actions from raw traffic data, the LLM receives a structured summary containing the current traffic state, predicted future traffic state, candidate action generated by the LSTM-based controller, and relevant operational constraints.

The input to the LLM is formatted as a structured JSON object:

```
{
  "intersection_id": "I1",
  "current_phase": "NS_Green",
  "elapsed_green_time": 25,
```

```
  "current_state": {
    "north_queue": 16,
    "south_queue": 18,
    "east_queue": 8,
    "west_queue": 7,
    "north_waiting_time": 76,
    "south_waiting_time": 82,
    "east_waiting_time": 29,
    "west_waiting_time": 25
  },
  "predicted_state_next_60s": {
    "north_queue": 24,
    "south_queue": 27,
    "east_queue": 10,
    "west_queue": 9,
    "north_waiting_time": 103,
    "south_waiting_time": 110,
    "east_waiting_time": 38,
    "west_waiting_time": 34
  },
  "candidate_action_from_lstm": "Extend_NS_Green",
  "constraints": {
    "min_green": 10,
    "max_green": 45,
    "yellow_time": 3,
    "pedestrian_request": false,
    "emergency_vehicle": false
  }
}
```

The LLM is instructed to generate a structured output rather than free-form text. The expected output format is:

```
{
  "accept_candidate_action": true,
  "recommended_action": "Extend_NS_Green",
  "recommended_duration": 10,
  "congestion_diagnosis": "The north-south approaches show increasing queue length and waiting time.",
  "risk_flag": "low",
  "safety_check": "No constraint violation detected.",
  "explanation": "The predicted queue and waiting time on the north-south approaches are significantly higher than those on the east-west approaches. Extending the current phase can reduce near-future congestion without violating the maximum green constraint."
}
```

The LLM module has four major functions.

First, it evaluates whether the candidate action generated by the LSTM-based controller is reasonable under the predicted traffic state. Second, it diagnoses traffic patterns, such as directional congestion, sudden demand surge, or potential spillback. Third, it recommends whether to accept, reject, or adjust the candidate action. Fourth, it produces a natural-language explanation to support human interpretability.

The LLM output is represented as:

$$r_t^{LLM} = [g_t, d_t, z_t, e_t],$$

where $g_t$ is the recommended action, $d_t$ is the recommended duration, $z_t$ is the congestion or risk diagnosis, and $e_t$ is the explanation.

## 2.7. Safety-Constrained Action Filtering

Because traffic signal control is safety-critical, the LLM recommendation cannot be directly executed. A safety filter is introduced to ensure that all final actions satisfy traffic signal operation constraints.
The constraint set is defined as:
$$C = \{c_1, c_2, \ldots, c_m\},$$

where each $c_i$ represents a traffic control constraint. In this study, the main constraints include:
$$T_{green} \geq T_{min},$$
$$T_{green} \leq T_{max},$$
$$T_{yellow} = T_{yellow}^{req},$$

and pedestrian or emergency vehicle constraints:
$$a_t \in A_{safe}.$$

Here, $T_{green}$ is the green time of the current phase, $T_{min}$ is the minimum green time, $T_{max}$ is the maximum green time, and $T_{yellow}^{req}$ is the required yellow interval. $A_{safe}$ denotes the set of signal actions that are allowed under pedestrian crossing or emergency vehicle conditions.
The final executable action is obtained by applying the safety filter:
$$a_t^{final} = S(a_t^{LSTM}, r_t^{LLM}, C),$$

where $a_t^{LSTM}$ is the candidate action generated by the LSTM-based predictive controller, $r_t^{LLM}$ is the LLM recommendation, $C$ is the constraint set, and $S(\cdot)$ is the safety filtering function.
If the LLM recommendation satisfies all constraints, the recommendation is accepted. If the recommendation violates any constraint, the system either modifies it to the nearest feasible action or falls back to the original LSTM-based candidate action. For example, if the LLM recommends extending the green phase beyond the maximum green duration, the safety filter limits the extension to the maximum allowable duration. If a pedestrian request is active, the filter prevents phase skipping that would violate pedestrian crossing requirements.
This design ensures that the LLM can improve interpretability and adaptive reasoning without compromising operational safety.

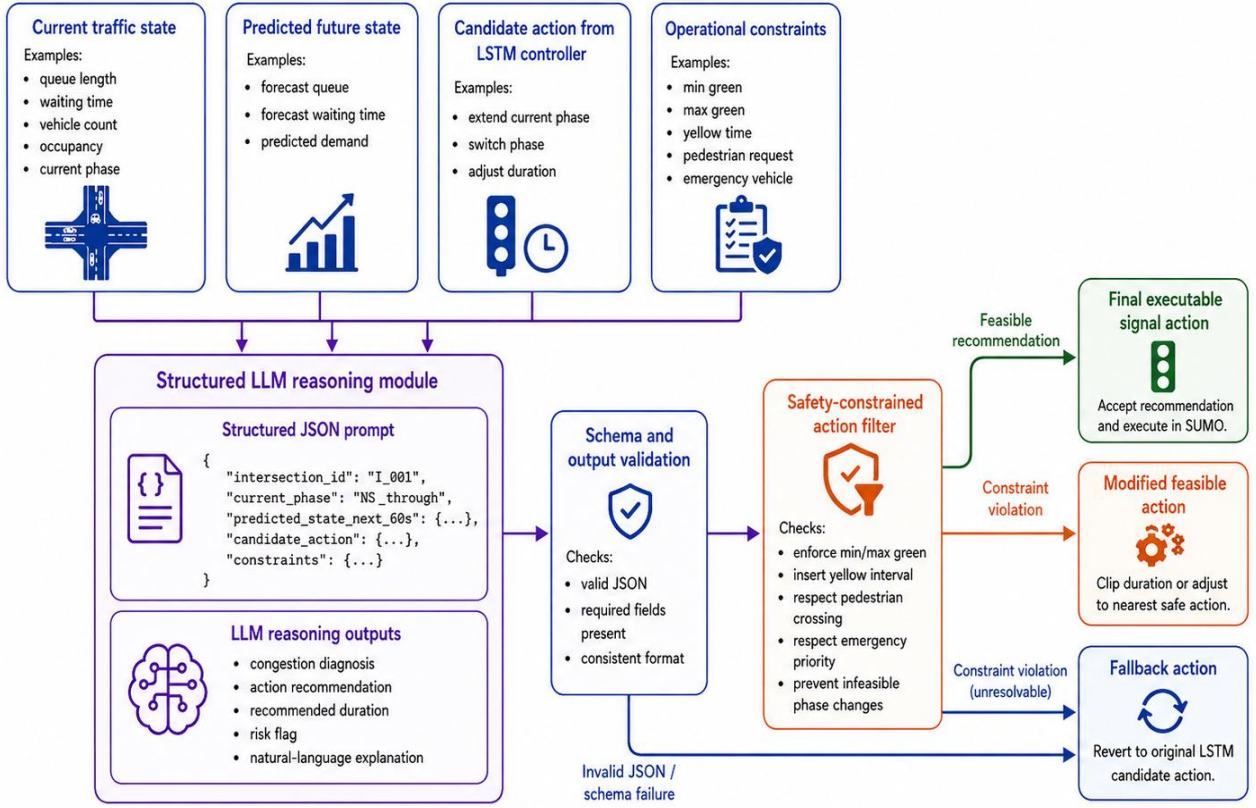

Figure 4. Structured LLM reasoning and safety-constrained action filtering process

**2.8. Proposed LLM-Augmented Traffic Signal Control Algorithm**
The complete control procedure is summarized as follows.
**Algorithm 1: LLM-Augmented Traffic Signal Control**
**Input:**
Historical traffic states $X_t$, feasible phase set $A$, traffic signal constraints $C$, trained LSTM model $f_\theta$, LLM decision module $\mathcal{M}_{LLM}$
**Output:**
Final signal action $a_t^{final}$

1. Collect current traffic state $s_t$ from the simulation environment.
2. Construct the historical state sequence:
$$X_t = [s_{t-K+1}, \ldots, s_t].$$

3. Use the LSTM model to predict future traffic states:
$$\hat{Y}_{t+1:t+H} = f_\theta(X_t).$$

4. Compute the predicted demand score $D(a)$ for each candidate phase $a \in A$.
5. Generate the LSTM-based candidate action:
$$a_t^{LSTM} = \arg\max_{a \in A} D(a).$$

6. Construct a structured LLM input using the current state, predicted state, candidate action, and signal constraints.
7. Obtain the LLM recommendation:
$$r_t^{LLM} = \mathcal{M}_{LLM}(s_t, \hat{Y}_{t+1:t+H}, a_t^{LSTM}, C).$$

8. Apply the safety filter:
$$a_t^{final} = S(a_t^{LSTM}, r_t^{LLM}, C).$$
9. Execute $a_t^{final}$ in the traffic simulation environment.
10. Record evaluation metrics, including delay, queue length, travel time, throughput, number of stops, and emissions.

This algorithm is repeated at every signal decision interval until the simulation ends.

## 2.9. Baseline Methods
To evaluate the effectiveness of the proposed framework, we compare it with several baseline methods.
### 2.9.1. Fixed-Time Control
The fixed-time controller applies a predefined signal cycle and phase duration regardless of real-time traffic conditions. This method is simple and widely used as a basic benchmark.
### 2.9.2. Rule-Based Actuated Control
The rule-based controller adjusts the signal phase according to current queue length or waiting time. If the queue length of the current green phase remains above a threshold, the green phase is extended. Otherwise, the controller switches to another phase.
### 2.9.3. LSTM-Based Predictive Control
The LSTM-based predictive controller uses the trained LSTM model to forecast future traffic states and selects the phase with the highest predicted demand score. This baseline is the most important comparison because it isolates the contribution of the LLM module.
### 2.9.4. Proposed LLM-Augmented LSTM Control
The proposed method uses the LSTM-based predictive controller to generate candidate actions and then applies LLM-based reasoning and safety-constrained filtering to produce the final signal action.

## 2.10. Implementation Design
The proposed framework is implemented in a SUMO-based microscopic traffic simulation environment. Traffic state variables are extracted at fixed intervals through the TraCI interface. The LSTM model is trained offline using simulation-generated traffic trajectories under multiple demand patterns. During online control, the trained LSTM model predicts near-future traffic states, while the LLM module is called at each signal decision interval or only when abnormal traffic conditions are detected.

To reduce computational cost and improve stability, the LLM module is not required to operate at every simulation step. Instead, it can be triggered under one of the following conditions:
$$\Delta q_t > \tau_q,$$
$$\Delta w_t > \tau_w,$$

or

$$D(a_1) - D(a_2) < \tau_D,$$

where $\Delta q_t$ is the recent increase in queue length, $\Delta w_t$ is the recent increase in waiting time, and $D(a_1) - D(a_2)$ is the difference between the top two candidate phase demand scores. This trigger-based design allows the LLM to focus on ambiguous, abnormal, or high-impact decision points rather than routine control cases.

The LLM temperature is set to a low value to improve output consistency. All LLM responses are required to follow a predefined JSON schema. Responses that fail schema validation are rejected and replaced by the original LSTM-based candidate action.

# 3. SIMULATION SETUP

## 3.1. Simulation Environment
The experiments are conducted using Simulation of Urban MObility (SUMO), a microscopic traffic simulator widely used in intelligent transportation research. The proposed framework interacts with SUMO through the TraCI interface, which enables real-time traffic state extraction and dynamic signal phase modification.

The simulated network is a four-arm urban intersection. Each approach contains two incoming lanes, including one through lane and one left-turn/shared lane. The intersection contains four signal phases:

$$A = \{NS\_Green, EW\_Green, NS\_Left, EW\_Left\}.$$

Each phase is followed by a 3-second yellow interval. The minimum green time is set to 10 seconds, and the maximum green time is set to 45 seconds. Each simulation run lasts 3600 seconds. To reduce randomness, each experiment is repeated five times using different random seeds, and the average results are reported.

Traffic states are extracted every 10 seconds. At each decision step, the controller receives queue length, waiting time, vehicle count, average speed, lane occupancy, and current signal phase.

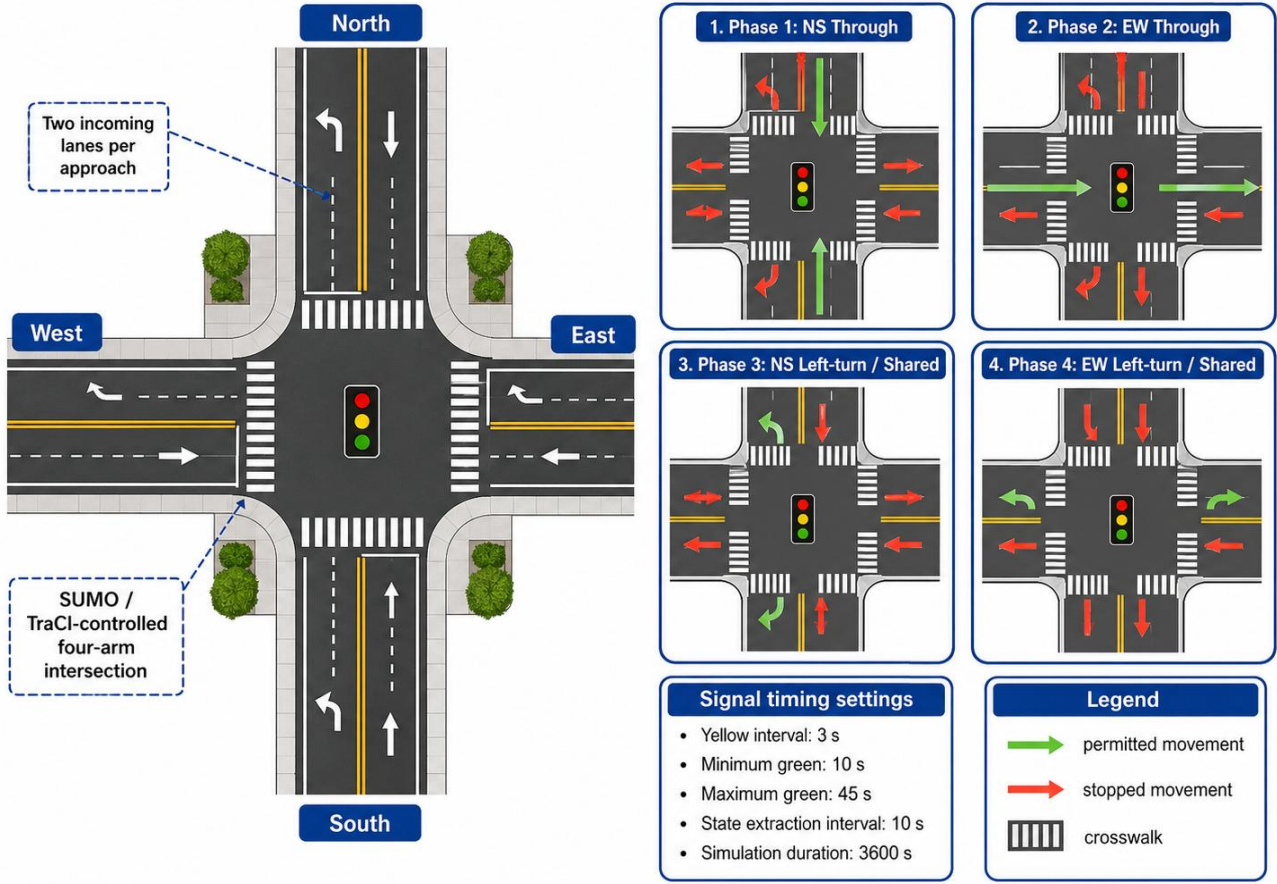

Figure 5. SUMO simulation network and signal phase design

### 3.2. Traffic Demand Scenarios
To evaluate the robustness of different control strategies, three traffic demand scenarios are designed.
### 3.2.1. Balanced Demand
In the balanced demand scenario, vehicles arrive from all four approaches with similar flow rates. This scenario represents a normal urban intersection with relatively stable demand. The arrival rate is set to approximately 450 vehicles per hour per approach.
### 3.2.2. Directional Peak Demand
In the directional peak scenario, the north-south direction has significantly higher demand than the east-west direction. This scenario represents a typical morning or evening peak-hour pattern. The north-south approaches have an arrival rate of approximately 700 vehicles per hour per approach, while the east-west approaches have an arrival rate of approximately 350 vehicles per hour per approach.
### 3.2.3. Sudden Surge Demand
In the sudden surge scenario, traffic demand is initially balanced. After 1200 seconds, the northbound and southbound arrival rates increase sharply for 900 seconds. This scenario simulates non-recurrent traffic conditions, such as temporary demand surges caused by events, incidents, or upstream congestion propagation.

### 3.3. Baseline Methods
The proposed method is compared with four baseline methods.
#### 3.3.1. Fixed-Time Control
The fixed-time controller uses a predefined signal cycle and phase duration. It does not respond to real-time traffic states. Each major through phase is assigned a 30-second green time, while each protected turn phase is assigned a 15-second green time.
#### 3.3.2. Rule-Based Actuated Control
The rule-based controller adjusts phase duration according to the current queue length and waiting time. If the queue length of the current green phase remains above a predefined threshold, the green phase is extended. Otherwise, the controller switches to the next phase.
#### 3.3.3. LSTM-Based Predictive Control
The LSTM-based predictive controller uses historical traffic states to forecast future queue length, waiting time, vehicle count, and occupancy. Based on the predicted future state, it calculates the demand score for each candidate phase and selects the phase with the highest predicted demand.
#### 3.3.4. LLM-Augmented LSTM Control
The proposed method first uses the LSTM-based predictive controller to generate a candidate signal action. The LLM module then receives the current traffic state, predicted traffic state, candidate action, and signal constraints. It provides a structured recommendation, congestion diagnosis, and explanation. A safety filter validates the recommendation before execution.

### 3.4. LSTM Training Setup
The LSTM model is trained using simulation-generated traffic trajectories collected from multiple demand patterns, including balanced demand, directional peak demand, and fluctuating demand. The input sequence length is set to 12 time steps, corresponding to 120 seconds of historical observations. The prediction horizon is set to 6 time steps, corresponding to 60 seconds of future traffic states.
The input features include:
$$[q_t, w_t, n_t, v_t, o_t, p_t],$$

where $q_t$ is queue length, $w_t$ is waiting time, $n_t$ is vehicle count, $v_t$ is average speed, $o_t$ is lane occupancy, and $p_t$ is the one-hot encoded signal phase.
The LSTM model contains two recurrent layers with 64 hidden units each. A fully connected layer is used to output the predicted traffic state. The model is trained using the Adam optimizer with a learning rate of 0.001. The batch size is 64, and the model is trained for 100 epochs. Mean squared error is used as the prediction loss. The training, validation, and testing split is 70%, 15%, and 15%, respectively.

### 3.5. LLM Decision Module Setup
The LLM module is implemented as a structured decision-support component. At each decision interval, the module receives a JSON-formatted input containing the current traffic state, predicted traffic state, LSTM-generated candidate action, and operational constraints.
To improve stability, the LLM temperature is set to 0. The output is required to follow a predefined JSON schema containing the recommended action, recommended duration, congestion diagnosis, risk flag, safety check result, and natural-language explanation.
If the LLM output fails schema validation, the system discards the LLM recommendation and falls back to the LSTM-based candidate action. All recommended actions are further checked by the safety filter before execution.
The LLM module is triggered only under selected conditions rather than at every simulation step. Specifically, it is activated when recent queue length or waiting time increases significantly, or when the demand-score difference between the top two candidate phases is small. This design reduces unnecessary LLM calls and allows the LLM to focus on abnormal or ambiguous traffic conditions.

### 3.6. Evaluation Metrics
The following metrics are used to evaluate control performance:
- **Average waiting time:** the average time vehicles spend waiting at the intersection.
- **Average queue length:** the average number of queued vehicles across all approaches.

- **Average travel time:** the average time required for vehicles to pass through the network.
- **Throughput:** the total number of vehicles that successfully leave the network.
- **Number of stops:** the average number of stops per vehicle.
- **$CO_2$ emission:** the estimated vehicle emission during the simulation.
- **Constraint violation rate:** the percentage of signal decisions violating operational constraints.
- **LLM explanation validity rate:** the percentage of LLM explanations consistent with the traffic state and final action.

For traffic efficiency metrics, lower values are better except for throughput, where higher values indicate better performance.

## 4. EXPERIMENTS AND RESULTS

The following results are illustrative simulated results used to demonstrate the expected analysis format. They should be replaced with actual SUMO simulation outputs in the final version.

### 4.1. Overall Performance under Different Demand Scenarios

Table 1 reports the overall performance comparison across the three traffic demand scenarios.

Table 1. Overall performance comparison across different demand scenarios.

| Scenario | Method | Avg. Waiting Time (s) | Avg. Queue Length | Avg. Travel Time (s) | Stops / Vehicle | Throughput | $CO_2$ Emission (kg) |
|---|---|---|---|---|---|---|---|
| Balanced | Fixed-Time | 42.6 | 13.8 | 118.4 | 2.31 | 1742 | 96.5 |
| Balanced | Rule-Based | 36.9 | 11.7 | 109.2 | 2.05 | 1788 | 90.3 |
| Balanced | LSTM-Predictive | 31.4 | 9.8 | 101.6 | 1.82 | 1835 | 84.7 |
| Balanced | LLM-Augmented LSTM | 29.8 | 9.1 | 98.7 | 1.71 | 1852 | 82.9 |
| Directional Peak | Fixed-Time | 68.3 | 24.6 | 156.8 | 3.12 | 2014 | 142.7 |
| Directional Peak | Rule-Based | 57.5 | 20.9 | 141.3 | 2.78 | 2096 | 132.4 |
| Directional Peak | LSTM-Predictive | 48.2 | 17.1 | 129.5 | 2.41 | 2168 | 121.9 |
| Directional Peak | LLM-Augmented LSTM | 43.6 | 15.2 | 122.7 | 2.19 | 2215 | 116.3 |
| Sudden Surge | Fixed-Time | 82.7 | 31.5 | 181.6 | 3.74 | 1958 | 168.9 |
| Sudden Surge | Rule-Based | 70.4 | 27.2 | 164.3 | 3.31 | 2041 | 154.8 |
| Sudden Surge | LSTM-Predictive | 61.8 | 23.4 | 151.9 | 2.96 | 2117 | 143.6 |
| Sudden Surge | LLM-Augmented LSTM | 52.9 | 19.1 | 138.5 | 2.53 | 2194 | 132.2 |

The results show that the LSTM-based predictive controller consistently outperforms fixed-time and rule-based control across all demand scenarios. This indicates that short-term traffic forecasting provides useful information for adaptive signal control.

The proposed LLM-augmented method further improves performance over the LSTM baseline. The improvement is relatively modest under balanced demand but becomes more significant under directional peak and sudden surge scenarios. This pattern suggests that the LLM module is most useful when traffic conditions are irregular, asymmetric, or rapidly changing.

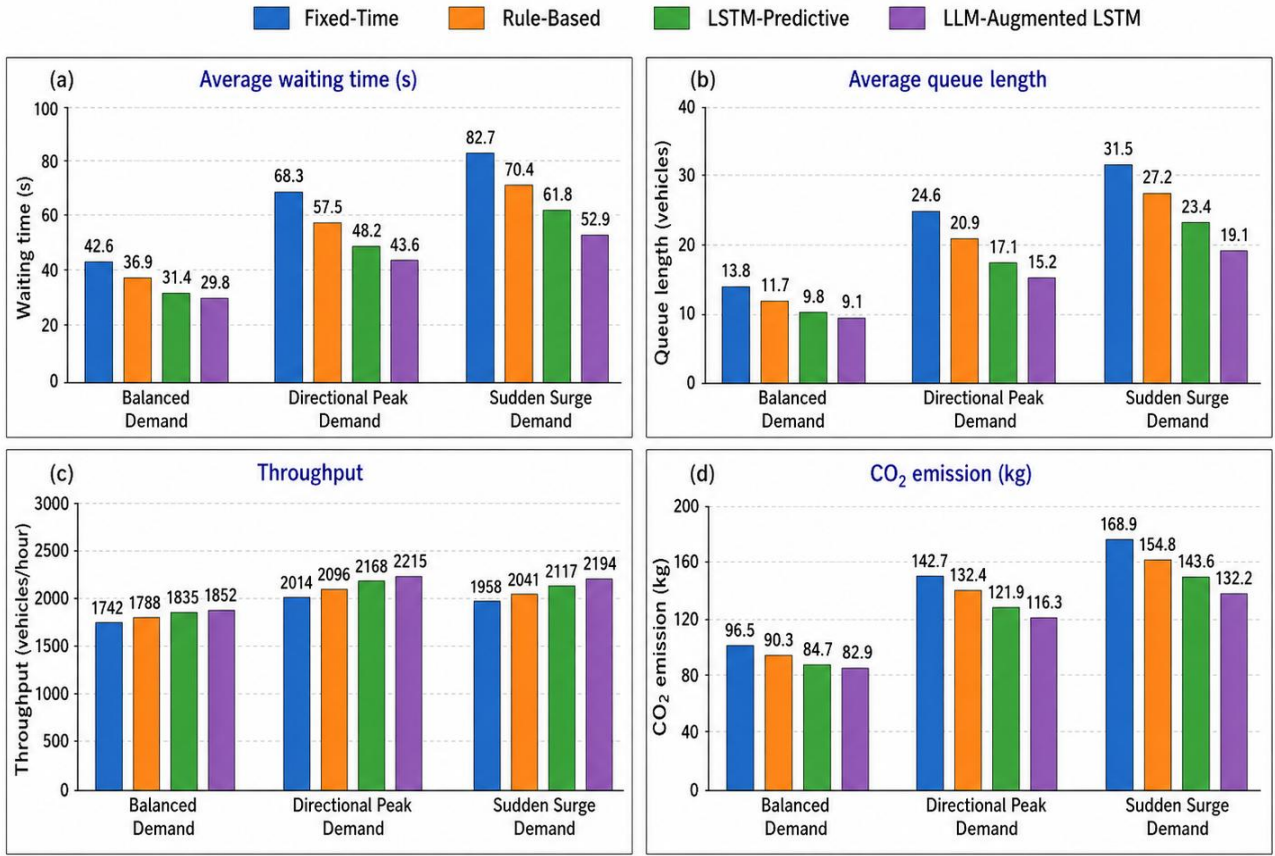

Figure 6. Performance comparison under different traffic demand scenarios

## 4.2. Relative Improvement over LSTM Baseline

Table 2 summarizes the relative improvement of the proposed method compared with the LSTM-based predictive controller. The improvement is smallest under balanced demand and largest under sudden surge demand. In routine traffic conditions, the LSTM predictive controller already performs well, leaving limited room for improvement. In contrast, under sudden demand changes, the LLM module provides additional contextual reasoning and supports more robust signal adjustment.

Table 2. Relative improvement of LLM-Augmented LSTM over LSTM-Predictive baseline.

| Scenario | Waiting Time Reduction | Queue Length Reduction | Travel Time Reduction | $CO_2$ Reduction | Throughput Increase |
|---|---|---|---|---|---|
| Balanced Demand | 5.1% | 7.1% | 2.9% | 2.1% | 0.9% |
| Directional Peak Demand | 9.5% | 11.1% | 5.3% | 4.6% | 2.2% |
| Sudden Surge Demand | 14.4% | 18.4% | 8.8% | 7.9% | 3.6% |

## 4.3. Ablation Study

To further analyze the contribution of each component, an ablation study is conducted under the sudden surge scenario. Four variants are compared:
- Full model: LSTM prediction + LLM reasoning + safety filter.
- Without LLM: equivalent to the LSTM-based predictive controller.
- Without safety filter: the LLM recommendation is directly applied if valid JSON is produced.
- Without prediction: the LLM only receives the current traffic state without LSTM-predicted future states.

Table 3. Ablation study under sudden surge demand.

| Method Variant | Avg. Waiting Time (s) | Avg. Queue Length | Avg. Travel Time (s) | Constraint Violation Rate |
|---|---|---|---|---|
| Full Model | 52.9 | 19.1 | 138.5 | 0.0% |
| Without LLM | 61.8 | 23.4 | 151.9 | 0.0% |
| Without Safety Filter | 50.7 | 18.6 | 136.2 | 4.8% |
| Without Prediction | 58.6 | 21.7 | 147.8 | 0.0% |

The ablation results show that both the LSTM prediction module and the LLM decision-support module contribute to the final performance. Removing the LLM module leads to a clear performance drop, indicating that the LLM provides additional reasoning beyond predictive control alone. Removing the prediction module also degrades performance, suggesting that the LLM benefits from future traffic state estimates rather than relying only on current observations. The variant without the safety filter achieves slightly lower waiting time, but it introduces constraint violations. This result confirms the necessity of the safety filter in safety-critical traffic signal control.

### 4.4. LLM Decision Analysis

The LLM module is triggered more frequently under directional peak and sudden surge scenarios, indicating that the trigger mechanism successfully identifies more complex or abnormal traffic conditions. The action adjustment rate also increases under sudden surge demand, suggesting that the LLM module is more active when the LSTM baseline faces more uncertain or unstable traffic states.

The explanation validity rate remains above 89% in all scenarios, indicating that most LLM explanations are consistent with the observed traffic states and final signal actions. After safety filtering, the constraint violation rate remains 0% across all scenarios.

Table 4 summarizes the behavior of the LLM module under different traffic scenarios.

| Scenario | LLM Trigger Rate | Candidate Action Acceptance Rate | Action Adjustment Rate | Explanation Validity Rate | Constraint Violation after Filtering |
|---|---|---|---|---|---|
| Balanced Demand | 18.6% | 84.2% | 15.8% | 93.5% | 0.0% |
| Directional Peak Demand | 31.4% | 76.8% | 23.2% | 91.7% | 0.0% |
| Sudden Surge Demand | 46.9% | 68.5% | 31.5% | 89.6% | 0.0% |

### 4.6. Discussion

The experimental results suggest three major findings.

First, LSTM-based prediction improves signal control compared with fixed-time and rule-based methods. This confirms that short-term traffic forecasting is useful for adaptive signal control because it allows the controller to respond not only to current congestion but also to near-future traffic evolution.

Second, the proposed LLM-augmented framework provides additional performance improvement over the LSTM baseline, especially under directional peak and sudden surge demand. The improvement is relatively small under balanced demand but becomes more significant as traffic conditions become more dynamic and abnormal. This supports the design motivation that LLM reasoning is most useful when the traffic state is ambiguous, non-recurrent, or difficult to handle using purely numerical rules.

Third, the safety filter is necessary for practical deployment. Although unconstrained LLM recommendations may sometimes produce more aggressive control actions, they can also violate traffic signal timing constraints. The proposed safety-constrained design prevents such unsafe decisions while preserving the interpretability benefits of the LLM module.

Overall, the results demonstrate that the proposed framework can combine the temporal prediction ability of LSTM, the contextual reasoning ability of LLMs, and the operational reliability of safety-constrained control. This layered design provides a practical path for incorporating LLMs into intelligent transportation systems without relying on unconstrained LLM outputs for direct real-time signal control.

## 5. SUMMARY

This study proposed an LLM-augmented traffic signal control framework that integrates LSTM-based traffic state prediction, predictive phase selection, structured LLM reasoning, and safety-constrained action filtering. Instead of using the LLM as a direct real-time controller, the proposed framework positions it as a higher-level decision-support module that interprets predicted traffic conditions, diagnoses congestion patterns, and provides explainable control recommendations.

The LSTM module was used to forecast short-term traffic states, including queue length, waiting time, vehicle count, and lane occupancy. Based on these predictions, a predictive controller generated candidate signal actions. The LLM module then evaluated these candidate actions using structured traffic-state inputs and produced recommendations with natural-language explanations. To ensure operational reliability, all LLM-generated suggestions were passed through a safety filter before execution.

Simulation-based experiments demonstrated that the proposed method can improve traffic efficiency compared with fixed-time control, rule-based control, and the LSTM-based predictive baseline. The improvement was especially more noticeable under directional peak and sudden demand surge scenarios, where traffic conditions were more dynamic and abnormal. The ablation study further showed that both the LSTM prediction module and the LLM reasoning module contributed to the overall performance, while the safety filter was necessary to prevent constraint violations.

Overall, the results suggest that LLMs can be effectively incorporated into intelligent traffic signal control when used as constrained reasoning and decision-support components rather than unconstrained low-level controllers. This framework provides a practical direction for building more adaptive, interpretable, and safety-aware traffic management systems.